\documentclass[11pt,a4paper]{article}

\usepackage{times}
\usepackage{latexsym}

\usepackage[utf8]{inputenc} 
\usepackage[T1]{fontenc}    
\usepackage[english]{babel}
\usepackage{amssymb, amsmath}
\usepackage{url}            
\usepackage{relsize}        
\usepackage[skip=5pt]{caption}
\usepackage{booktabs}       
\usepackage{amsfonts}       
\usepackage{nicefrac}       
\usepackage{microtype}      
\usepackage[pdftex]{graphicx}
\usepackage{float}
\usepackage[numbers,sort,compress]{natbib}
\usepackage{hyperref}       
\usepackage[hyperref]{myacl}

\aclfinalcopy 

\restylefloat{figure}
\DeclareMathOperator*{\argmax}{arg\,max}

\title{Exploring text datasets by visualizing relevant words}

\author{Franziska Horn$^1$, Leila Arras$^2$, Gr{\'e}goire Montavon$^1$,\\ \textbf{Klaus-Robert M{\"u}ller$^{1,3}$, and Wojciech Samek$^2$}\\
  $^1$Machine Learning Group, Technische Universit\"at Berlin, Berlin, Germany \\
  $^2$Machine Learning Group, Fraunhofer Heinrich Hertz Institute, Berlin, Germany \\
  $^3$Department of Brain and Cognitive Engineering, Korea University, Seoul, Korea \\
  {\tt franziska.horn@campus.tu-berlin.de}}

\date{}

\begin{document}
\setlength{\intextsep}{9pt plus 3pt minus 2pt}
\setlength{\textfloatsep}{9pt plus 3pt minus 2pt}
\setlength{\floatsep}{5pt plus 3pt minus 2pt}
\setlength{\dblfloatsep}{7pt plus 3pt minus 2pt}
\setlength{\dbltextfloatsep}{12pt plus 3pt minus 2pt}

\maketitle

\begin{abstract}
When working with a new dataset, it is important to first explore and familiarize oneself with it, before applying any advanced machine learning algorithms. However, to the best of our knowledge, no tools exist that quickly and reliably give insight into the contents of a selection of documents with respect to what distinguishes them from other documents belonging to different categories. In this paper we propose to extract `relevant words' from a collection of texts, which summarize the contents of documents belonging to a certain class (or discovered cluster in the case of unlabeled datasets), and visualize them in word clouds to allow for a survey of salient features at a glance. We compare three methods for extracting relevant words and demonstrate the usefulness of the resulting word clouds by providing an overview of the classes contained in a dataset of scientific publications as well as by discovering trending topics from recent New York Times article snippets.
\end{abstract}

\section{Introduction}
To avoid surprises when working with a new dataset, it is always important to first explore it and get an overview of the contained samples. This helps in developing an intuition whether, for example, the classification task is particularly difficult or if there are any outliers that might need special attention. Yet, while there exists an abundance of algorithms developed for squeezing out the last percentages of accuracy in categorization tasks, little work has focused on facilitating the exploratory analysis of text datasets, especially with respect to the actual contents of documents belonging to different classes. 

One way of getting an overview of a dataset is by visualizing it as a two dimensional scatter plot using dimensionality reduction techniques such as kernel PCA \cite{scholkopf1998nonlinear} or t-SNE \cite{van2008visualizing}. While this can show the match between natural clusters and assigned class labels and reveal outliers, it does not give insight into the \emph{content} of the samples belonging to individual clusters and classes. While one can (and should) manually inspect a few individual samples of each class, even more useful is an overview created by aggregating the content of all samples belonging to one cluster or class.

To get an overview of the documents belonging to different groups in a text dataset, we propose to create a word cloud for each class displaying its salient features, i.e.~relevant words contained in the samples assigned to this class. To extract these relevant words for each class, we compare three approaches, namely 1)~aggregating the raw tf-idf features of all samples from one class \cite{heimerl2014word}, 2)~using \emph{layerwise relevance propagation} (LRP) to break down the classification score of a linear classifier and project it onto the input features and then aggregate these scores for all documents of one class \cite{arras2016explaining}, and 3)~computing a relevancy score for each word by comparing how often it occurs in samples of one class compared to all others.
Additionally, we show that highlighting these relevant words in individual documents can be helpful for understanding which features contributed most to a classification decision. This can reveal why individual samples were misclassified, thereby exposing biases in the training set \cite{bach2015pixel,LapCVPR16}.

For unlabeled text datasets, which are becoming more and more frequent, e.g.~in the form of large data dumps leaked to journalists, getting a quick overview of the contents is even more important. In this case, the documents can first be clustered using algorithms like DBSCAN \cite{ester1996dbscan} before extracting the relevant words for each cluster.

We demonstrate how the extracted relevant words summarize the contents of different classes in a dataset of scientific publications. Furthermore, by identifying relevant words in clusters of recent New York Times article snippets, trending topics can be revealed.

All tools discussed in this paper as well as code to replicate the experiments are available as an open source Python library.\footnote{\url{https://github.com/cod3licious/textcatvis}}

\subsection{Related work}
Identifying relevant words in text documents was traditionally limited to the area of feature selection, where different approaches were used to discard `irrelevant' features in an attempt to improve the classification performance by reducing noise as well as save computational resources \cite{forman2003extensive}. However, the primary objective here was not to identify words that best describe the documents belonging to certain classes, but to identify features that are particularly uninformative in the classification task and can be disregarded. Other work was focused on selecting keywords for individual documents, e.g.~based on tf-idf variants \cite{lee2008news} or by using classifiers \cite{hulth2003improved,zhang2006keyword}. Yet, while these keywords might provide adequate summaries of single documents, they do not necessarily overlap with keywords found for other documents in the class and therefore it is difficult to aggregate them to get an overview of the contents of one class. Current tools available for creating word clouds as a means of summarizing a (collection of) document(s) mostly rely on term frequencies (while ignoring stopwords), possibly combined with part-of-speech tagging and named entity recognition to identify words of interest \cite{heimerl2014word,mcnaught2010using}. As we will show later, an approach based on tf-idf features does not reliably identify words that distinguish documents of one class from documents belonging to other classes. In more recent work, relevant features were selected using layerwise relevance propagation (LRP) to trace a classifier's decision back to the samples' input features \cite{bach2015pixel}. This was successfully used to understand the classification decisions made by a convolutional neural network (CNN) trained on a text categorization task and to subsequently determine relevant features for individual classes by aggregating the LRP scores computed on the test samples \cite{arras2016explaining,arras2016relevant}. We will compare such an LRP approach in the following as well, however, as a CNN requires a lot of data and typically several hours for training, we are using a linear classifier instead to ensure the relevant words can be extracted within minutes.

\section{Methods}
To get a quick overview of a text dataset, we want to identify and visualize the `relevant words' occurring in the collection of texts. We define relevant words as some characteristic features of the documents, which distinguish them from other documents. As the first step in this process, the texts therefore have to be preprocessed and transformed into feature vectors (Section~\ref{subsec:features}). While relevant words are supposed to occur often in the documents of interest, they should also distinguish them from other documents. When analyzing a whole dataset it is therefore most revealing to look at individual classes and obtain the relevant words for each class, i.e.~find the features that distinguish one class from another. If the dataset is unlabeled, the texts have to be clustered first and then relevant words can be selected for each cluster (Section~\ref{subsec:clustering}). 

The relevant words for a class or cluster $c$ are identified by computing a relevancy score $r_c$ for every word $t_i$ (with $i=1...T$, where $T$ is the number of unique terms in the given vocabulary) and then the word clouds are created using the top ranking words. The easiest way to compute relevancy scores is to simply check for frequent features in a selection of documents (Section~\ref{subsec:reltfidf}). However, this does not necessarily produce features that additionally occur infrequently in other classes. To improve on this, a classifier can be trained, thereby producing a set of weights for each class that, when applied to a feature vector, result in a classification score indicating how likely it is that this document is from a certain class. By examining this weight vector and applying it to the feature vectors of documents from a certain class, we obtain the features most important for distinguishing one class of documents from others in the classification task (Section~\ref{subsec:relsvm}). Another possibility is to directly compute a score for each word indicating in how many documents of one class it occurs compared to other classes (Section~\ref{subsec:reldist}).

\subsection{Preprocessing \& Feature extraction} \label{subsec:features}
All $N$ texts in a dataset are preprocessed by lowercasing and removing non-alphanumeric characters. Then each text is transformed into a \emph{bag-of-words} (BOW) feature vector $\mathbf{x}_k \in \mathbb{R}^T \;\forall k \in 1...N$ by first computing a normalized count, the \emph{term frequency} (tf), for each word in the text, and then weighting this by the word's \emph{inverse document frequency} (idf) to reduce the influence of very frequent but inexpressive words that occur in almost all documents (such as `and' and `the')~\cite{irbook, yang1997comparative}. The idf of a term $t_i$ is calculated as the logarithm of the total number of documents, $|N|$, divided by the number of documents which contain term $t_i$, i.e.
\begin{align*}
 \text{idf}\,(t_i) &= \log {|N|\over |\{k \in N\text{ : }t_i \in k\}|}.
 \end{align*}
The entry corresponding to the term $t_i$ in the feature vector $\mathbf{x}_k$ of a document $k$ is then
\begin{align*}
\mathbf{x}_{ki} &= \text{tf}\,(t_i) \cdot \text{idf}\,(t_i).
\end{align*}
In addition to single terms, we are also considering meaningful combinations of two words (i.e.~bigrams) as features. However, to not inflate the feature space too much (since later, relevancy scores have to be computed for every feature), only distinctive bigrams are selected as detailed in Appendix \ref{sec:bigrams}.

\subsection{Clustering} \label{subsec:clustering}
For unlabeled datasets, the texts first have to be clustered to be able to extract relevant words for each cluster. For this, we use \emph{density-based spatial clustering of applications with noise} (DBSCAN) \cite{ester1996dbscan}, a clustering algorithm that identifies clusters as areas of high density in the feature space, separated by areas of low density. This algorithm was chosen as it does not assume that the clusters have a certain shape (unlike e.g.~the k-means algorithm, which assumes spherical clusters) and it allows for noise in the dataset, i.e. does not enforce that all samples belong to a certain cluster.

DBSCAN is based on pairwise distances between samples and first identifies `core samples' in areas of high density and then iteratively expands a cluster by joining them with other samples, whose distance is below some user defined threshold. As the cosine similarity is a reliable measure of similarity for text documents, we compute the pairwise distances used in the DBSCAN algorithm by first reducing the documents' tf-idf feature vectors to 250 linear kernel PCA components to remove noise and create more overlap between the feature vectors~\cite{scholkopf1998nonlinear}, and then compute the cosine similarity between these vectors and subtract it from $1$ to transform it into a distance measure. As clustering is an unsupervised process, a value for the distance threshold has to be chosen such that the obtained clusters seem reasonable. In the experiments described below, we found that a minimum cosine similarity of $0.55$ to other samples in the cluster (i.e.~using a distance threshold of $0.45$) leads to texts about the same topic being grouped together.

\subsection{Identifying relevant words} \label{subsec:relwords}
Relevant words for each class or cluster are identified by computing a relevancy score $r_c$ for every word $t_i$ based on the documents in the class or cluster and then selecting the highest scoring words. These scores can be computed either by aggregating the raw tf-idf features of all documents in the group (Section~\ref{subsec:reltfidf}), by aggregating these features weighted by some classifier's parameters (Section~\ref{subsec:relsvm}), or directly by computing a score for each word depending on the number of documents it occurs in from this class relative to other classes (Section~\ref{subsec:reldist}).

\subsubsection{Salient tf-idf features} \label{subsec:reltfidf}
A very simple and straightforward approach to identifying relevant words for one class or cluster $c$ is to simply add up the tf-idf feature vectors $\mathbf{x}_k$ of all documents belonging to this class or cluster, i.e.~where $y_k = c$. For one word $t_i$ this results in a relevancy score with respect to class $c$ given by
\begin{align*}
r_c\_\text{tfidf}\,(t_i) & = \sum_{k\,:\,y_k = c} \mathbf{x}_{ki}\,.
\end{align*}
The words with the highest scores occur in most documents of this class. These words also don't occur in most documents in the dataset, as otherwise their idf scores would have been close to zero. However, it is still possible that high scoring words occur equally often in documents of some other class, without this having dramatic effects on the idf weights. Therefore, while this score yields relevant words in the sense that they occur frequently in documents of the current class, these words are not necessarily distinguishing this class from others.

\subsubsection{Decomposed classifier scores (SVM+LRP)} \label{subsec:relsvm}
To find distinguishing words, we train a classifier to identify features based on which it can be decided to what class a sample belongs to. Linear classifiers have a weight vector $\mathbf{w}_c \in \mathbb{R}^T$ and bias term $b_c$ for each class $c$ and assign a new document to the class with the highest score after applying this weight vector to a document's tf-idf vector $\mathbf{x}_k$:
\begin{align*}
\hat y_k = \argmax_c \; b_c + \mathbf{w}_c^\top \mathbf{x}_k \,.
\end{align*}
As the tf-idf vectors are always positive, a large positive (or negative) weight $\mathbf{w}_{ci}$ indicates whether the corresponding word $t_i$ is providing evidence for (or against) this document belonging to class $c$. One idea might be to use these weights directly to identify relevant words. However, this could yield misleading results, since the classifier might have large weights for some words, which clearly identify a document as belonging to this class, but which only occur in very few documents and are therefore not representative for the class as a whole. Instead, the tf-idf vectors of all documents from one class are multiplied elementwise by the weight vector of this class and then summed up to yield the final relevance score for a term $t_i$ as
\begin{align*}
r_c\_\text{lrp}\,(t_i) & = \sum_{k\,:\,y_k = c} \left(\mathbf{w}_{ci} \mathbf{x}_{ki} + \frac{b_c}{T}\right)\,.
\end{align*}
This elementwise decomposition of the classification score is a special case of \emph{layerwise relevance propagation} (LRP) for a one-layer network. LRP was originally developed to better understand the classification process in deep neural networks (DNN) by propagating the classifier decision back to the input layer in order to visualize the features based on which the decision was made \cite{bach2015pixel,arras2016explaining,arras2016relevant,montavon2017methods}.

For the following experiments we are using a linear SVM \cite{muller2001introduction} to classify the documents, but other linear classifiers such as a logistic regression can be used as well. It is also possible to use a DNN together with LRP to compute the relevant words \cite{arras2016explaining,arras2017explaining}. However, DNN typically require a comparatively long time to train, while with a linear classifier it is possible to obtain the relevant words within minutes on a desktop computer.
It is important to note, however, that selecting relevant words with LRP only works if the classifier is fairly accurate, as otherwise the trained weights and resulting scores are meaningless. If a DNN clearly outperforms a linear classifier on a dataset, it should be used instead to identify relevant words.

\subsubsection{Distinctive words} \label{subsec:reldist}
\begin{figure}[!h]
  \centering
      \includegraphics[width=\columnwidth]{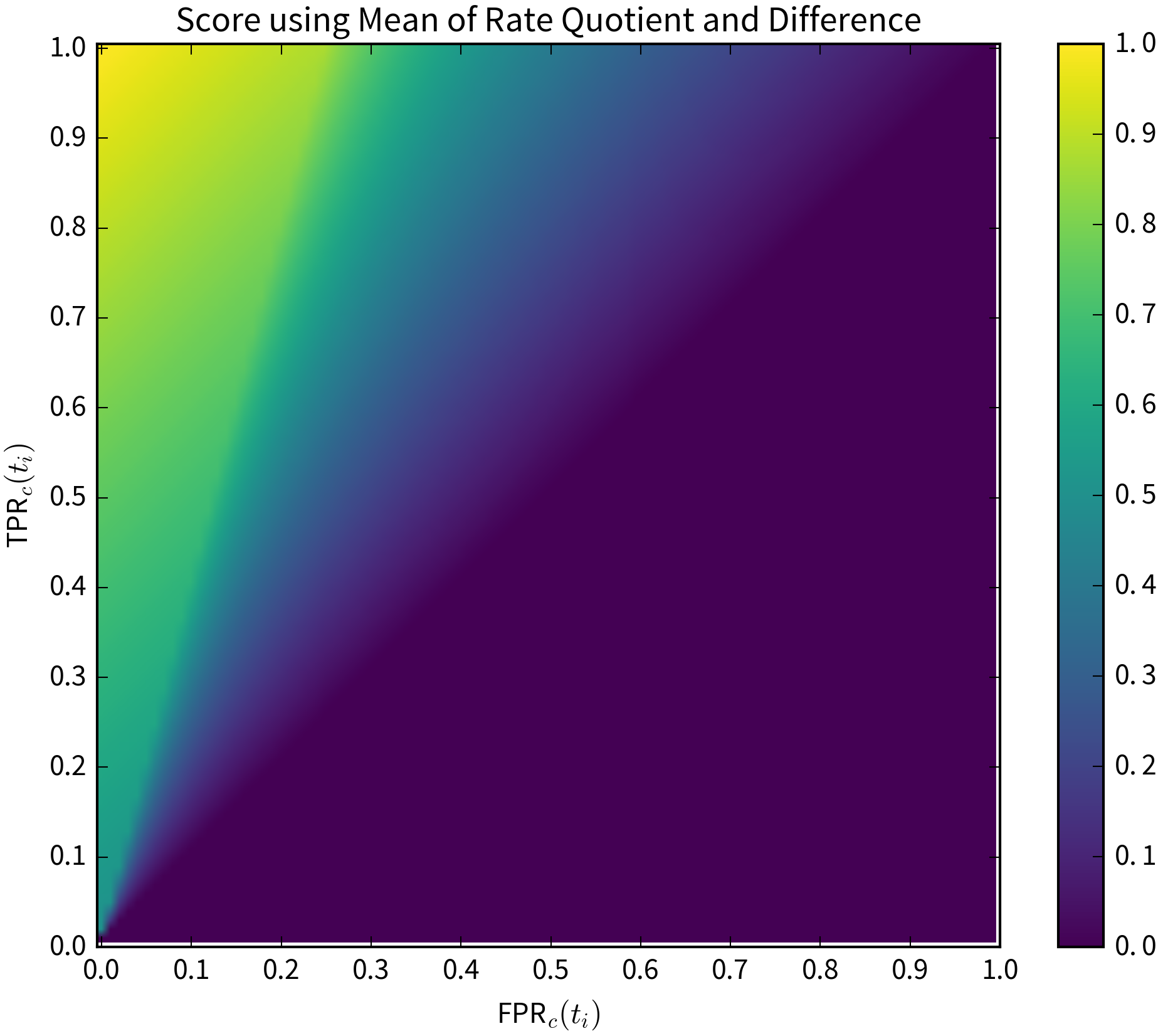}
  \caption{Score $r_c\_\text{dist}\,(t_i)$ depending on a word's TPR and FPR for one class.}
  \label{fig:distinctive_scores}
\end{figure}

Instead of aggregating (weighted) features of all documents, we can also compute a score directly for each word depending on the number of documents it appears in from a certain class compared to documents from other classes. We call the fraction of documents from a target class $c$ that contain the word $t_i$ this word's true positive rate
\begin{align*}
\text{TPR}_c(t_i) = {|\{k\text{ : }y_k = c \wedge \mathbf{x}_{ki} > 0\}|\over |\{k\text{ : }y_k = c\}|}\,.
\end{align*}
Correspondingly, we can compute a word's false positive rate as the mean plus the standard deviation of the TPRs of this word for all other classes:\footnote{We are not taking the maximum of the other classes' TPRs for this word to avoid a large influence of a class with maybe only a few samples, which can happen e.g.~when clustering the text documents.}
\begin{align*}
\text{FPR}_c(t_i) =& \text{ mean}(\{\text{TPR}_l(t_i) : l \neq c\})\\ &+ \text{std}(\{\text{TPR}_l(t_i) : l \neq c\})\,.
\end{align*}
The objective is to find words that occur in many documents from the target class (i.e.~have a large $\text{TPR}_c(t_i)$), but only occur in few documents of other classes (i.e.~have a low $\text{FPR}_c(t_i)$). One way to identify such words would be to compute the difference between both rates, i.e.
\begin{align*}
r_c\_\text{diff}\,(t_i) = \max\{\text{TPR}_c(t_i) - \text{FPR}_c(t_i), 0\}\,,
\end{align*}
which is similar to traditional feature selection approaches \cite{forman2003extensive}.
However, while this score yields words that occur more often in the target class than in other classes, it does not take into account the relative differences. For example, to be able to detect emerging topics in newspaper articles, we are not necessarily interested in words that occur often in today's articles and infrequently in yesterday's. Instead, we acknowledge that \emph{not most articles} published today will be written about some new event, only \emph{significantly more articles} compared to yesterday. Therefore, we propose instead a rate quotient, which gives a score of $1$ to every word that has a TPR about three times higher than its FPR:
\begin{align*}
r_c\_\text{quot}\,(t_i) &= {\min\{\max\{z_c(t_i),1\},4\}-1 \over 3},\\
 \text{with } z_c(t_i) &= {\text{TPR}_c(t_i) \over \max\{\text{FPR}_c(t_i), \epsilon\}}\,.
\end{align*}
While the rate quotient extracts relevant words that would otherwise go unnoticed, for a given FPR of $0.05$ it assigns the same score to words with a TPR of $0.3$ and a TPR of $1.0$. Therefore, to create a proper ranking amongst all relevant words, we take the mean of both scores to compute the final score,
\begin{align*}
r_c\_\text{dist}\,(t_i) & = 0.5\left(r_c\_\text{diff}\,(t_i) + r_c\_\text{quot}\,(t_i)\right),
\end{align*}
which results in the TPR-FPR relation shown in Fig.~\ref{fig:distinctive_scores}.

\section{Experiments \& Results}
To illustrate how the identified relevant words can help when exploring new datasets, we test the previously described methods on a corpus of scientific publications about various types of cancer as well as recent article snippets from the New York Times. The code to replicate the experiments is available online and includes functions to cluster documents, extract relevant words based on all three methods described above and visualize them in word clouds, as well as highlight relevant words in individual documents.\footnote{\url{https://github.com/cod3licious/textcatvis}}

\subsection{Cancer paper paragraphs}
\begin{figure}[!h]
  \centering
      \includegraphics[width=\columnwidth]{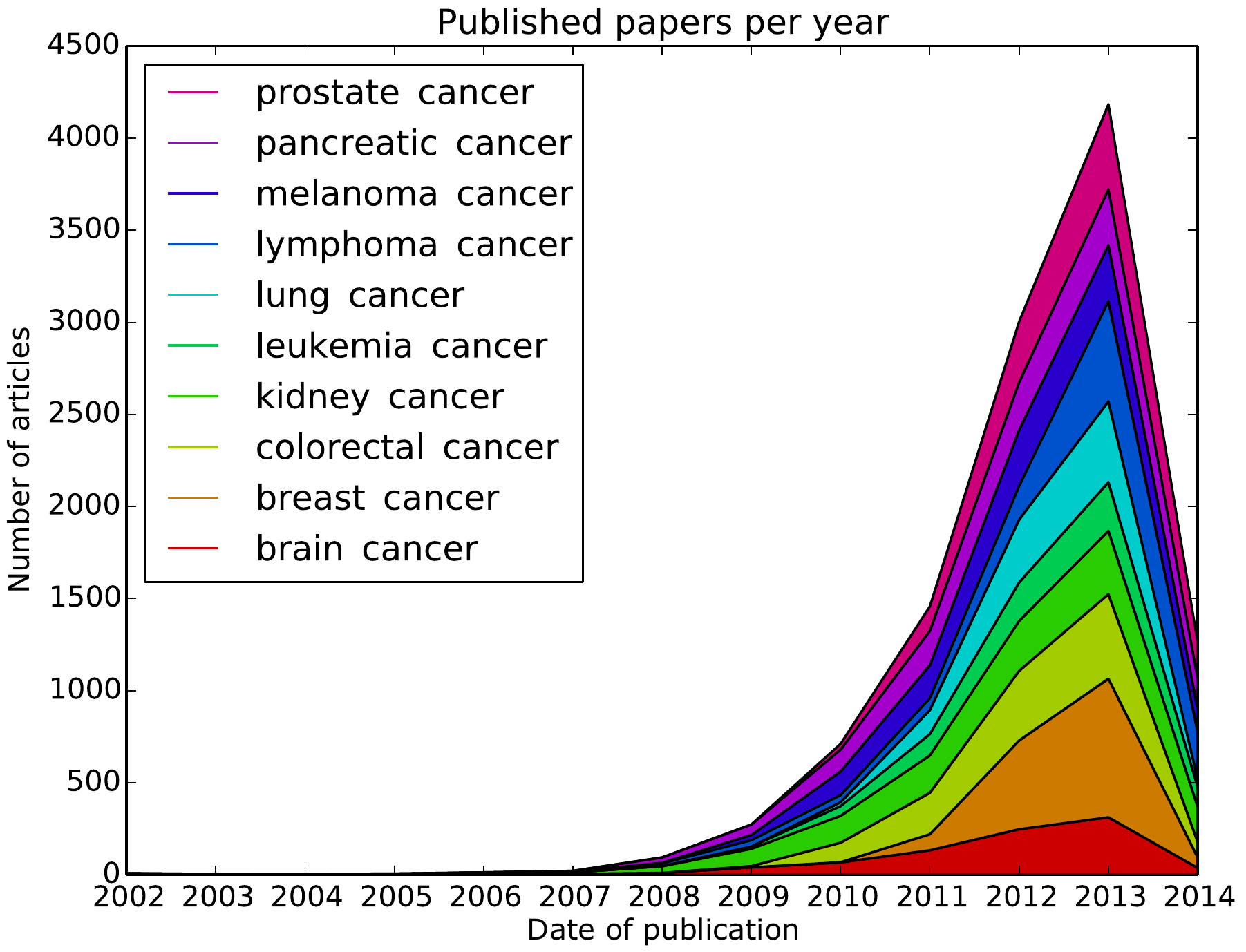}
  \caption{Number of full text cancer papers obtained for every cancer type by year. Note that the articles were downloaded in August 2014, which explains why there are significantly fewer papers for 2014 than the year before.}
  \label{fig:cancerdataset}
\end{figure}

We created a dataset of 11049 publicly available scientific publications by using the PubMed API\footnote{\url{http://www.ncbi.nlm.nih.gov/books/NBK25500/}} to download all full text PubMed Central papers associated with a keyword corresponding to one of ten different types of cancer (Fig.~\ref{fig:cancerdataset}). In a preprocessing step, the dataset was reduced to contain only articles (a)~associated with a single cancer type and (b)~reporting original research (i.e.~no editorials of journals, etc.). The paragraphs of each article were automatically assigned one of the paragraph type labels `Abstract', `Introduction', `Methods', `Results', `Discussion', or `Mixed', depending on the respective heading identified in the downloaded XML file. Ignoring the `mixed' paragraphs, we are left with almost 50k paragraphs from which we randomly subsampled 10k. Each of the paragraphs is associated with two labels: one for the paragraph type and another one for the cancer type the paper is about. The full dataset is available online.\footnote{\url{https://github.com/cod3licious/cancer_papers}}
\begin{figure}[!h]
  \centering 
      \includegraphics[width=\columnwidth]{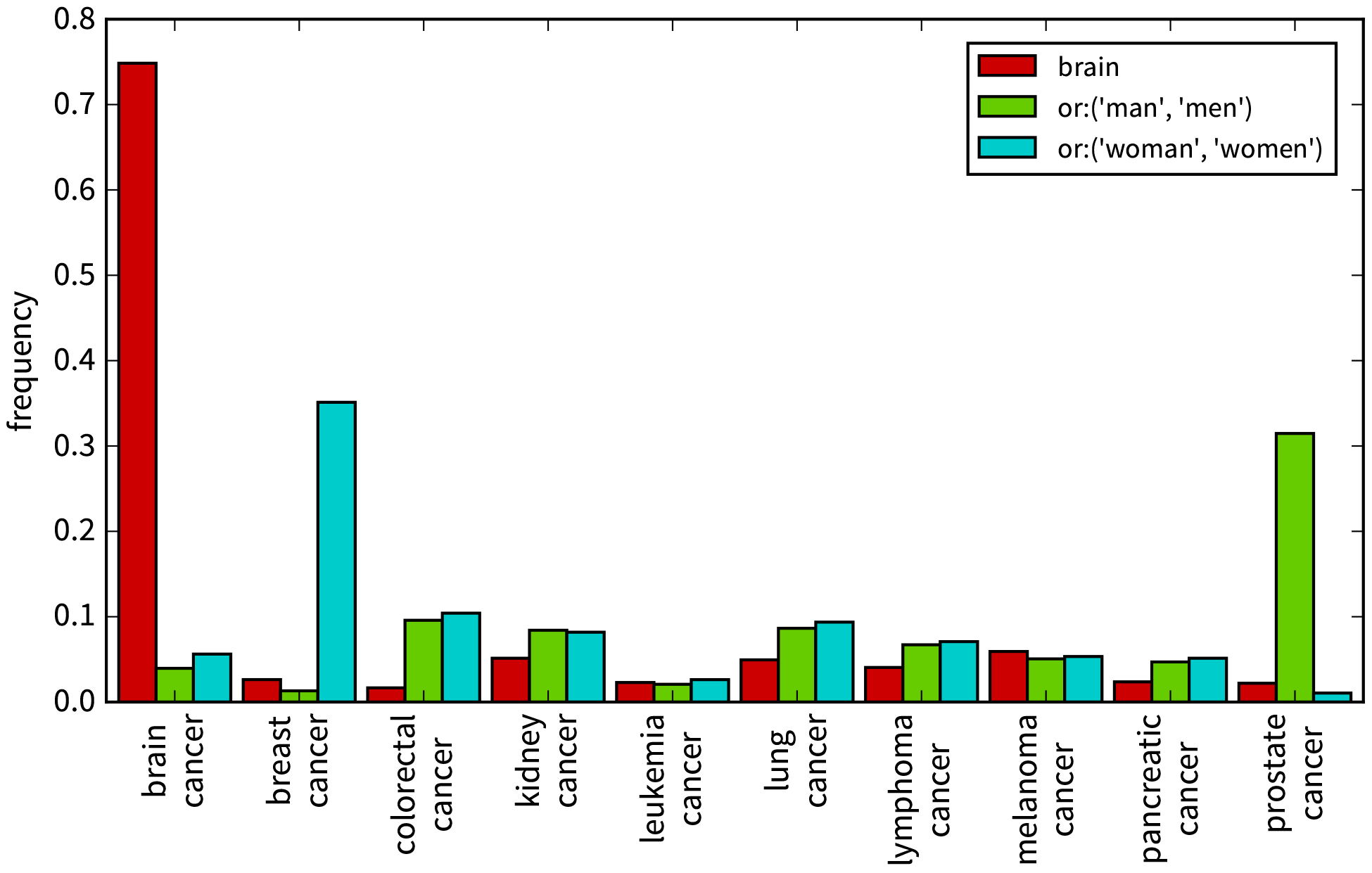}
  \caption{Fraction of texts in each category mentioning either `brain', `man'/`men', or `woman'/`women'.}
  \label{fig:cancer_occurrences}
\end{figure}
 
\begin{figure*}[!ht]
  \centering 
      \includegraphics[width=0.32\textwidth]{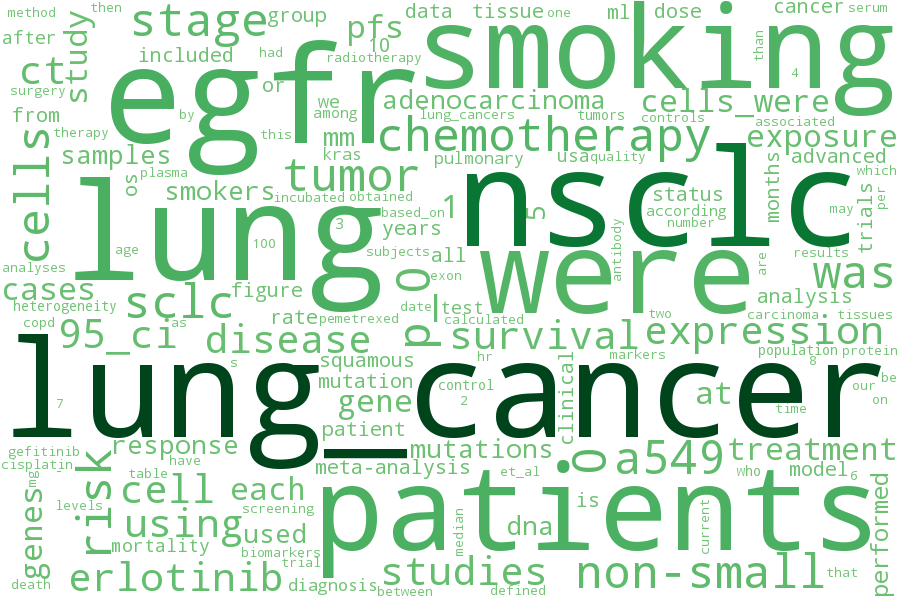}
      \includegraphics[width=0.32\textwidth]{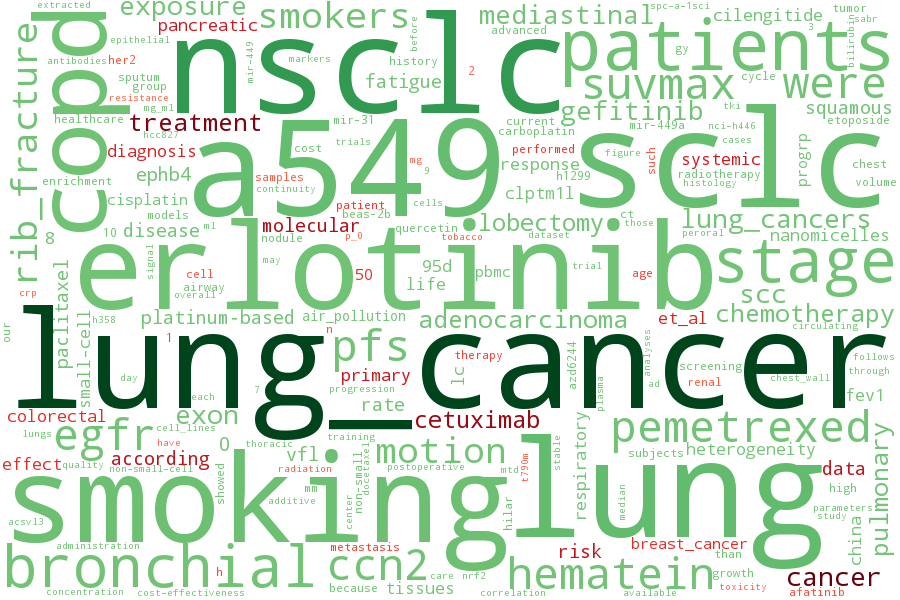}
      \includegraphics[width=0.32\textwidth]{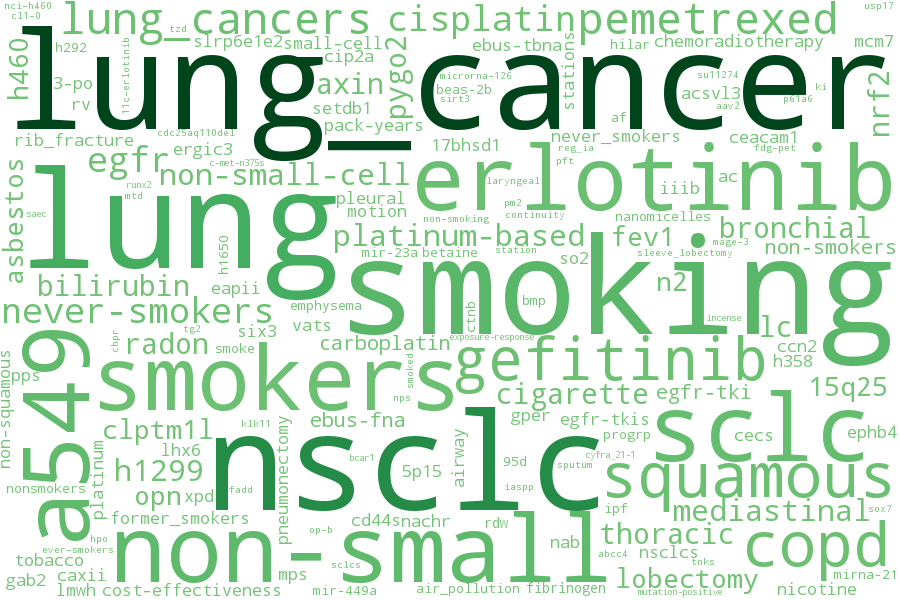}
  \caption{Word clouds created from the paragraphs of papers about lung cancer based on tf-idf features (\emph{left}), LRP of the classifier output from an SVM (\emph{middle}), and distinctive words (\emph{right}).}
  \label{fig:wordcloud_cancertype}
\end{figure*}
\begin{figure*}[!ht]
  \centering 
      \includegraphics[height=3.7cm]{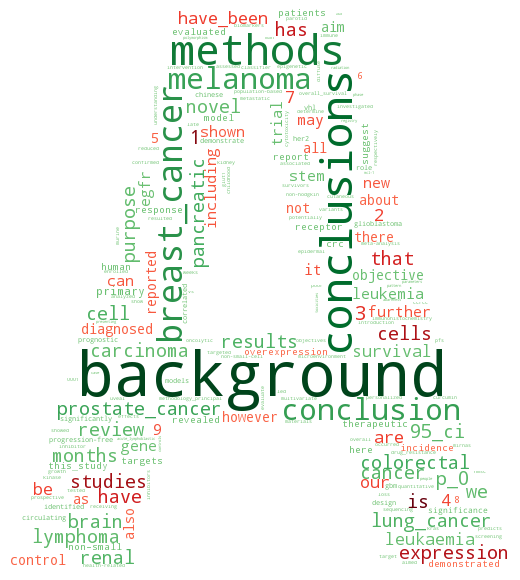}
      \includegraphics[height=3.7cm]{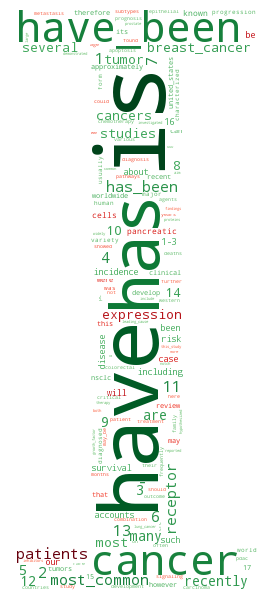}
      \includegraphics[height=3.7cm]{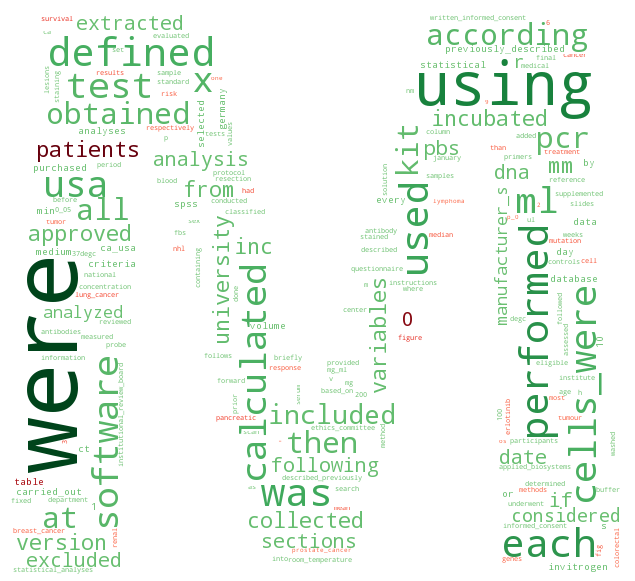}
      \includegraphics[height=3.7cm]{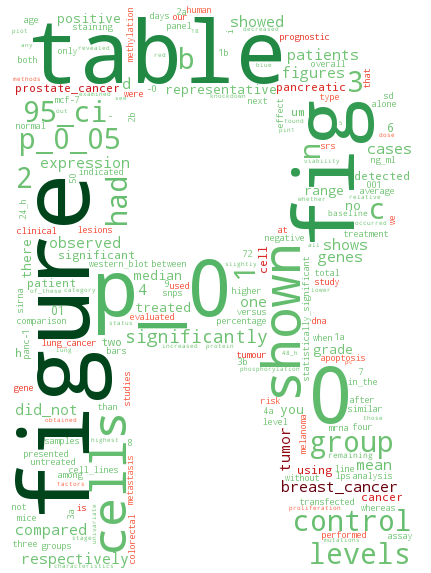}
      \includegraphics[height=3.7cm]{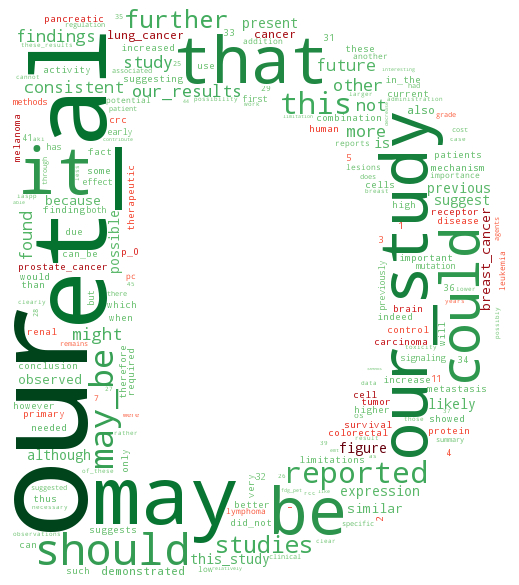}
  \caption{Word clouds created using LRP on the classifier output from an SVM trained to detect a text's paragraph type (\textbf{A}bstract, \textbf{I}ntroduction, \textbf{M}ethods, \textbf{R}esults, \textbf{D}iscussion).}
  \label{fig:wordcloud_partype}
\end{figure*}
\begin{figure*}[!ht]
  \centering \small
      \rotatebox{90}{\sf tf-idf features}
      ~
      \boxed{\includegraphics[width=0.8\textwidth,clip=True,trim=0 655 0 50]{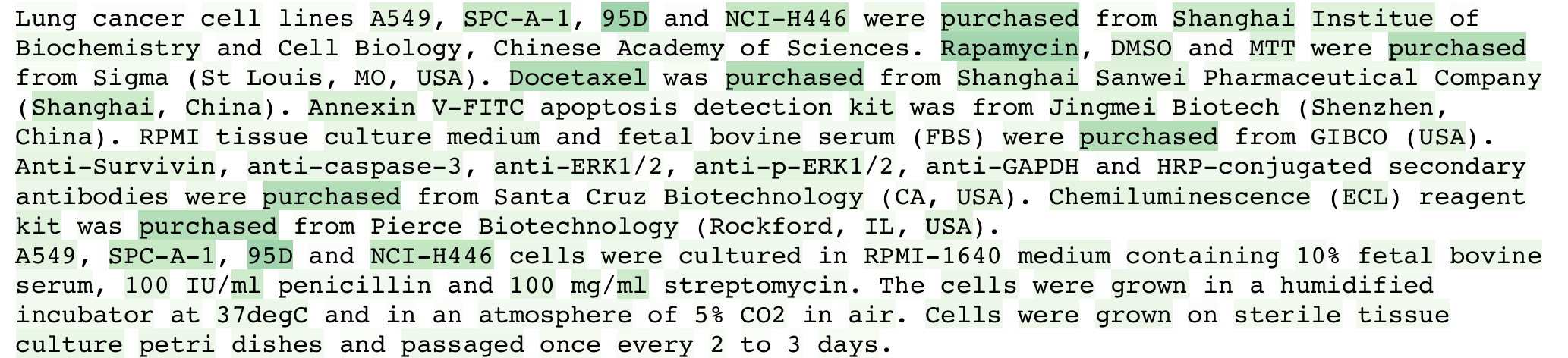}} \vskip 2mm
      \rotatebox{90}{\sf predicted \textbf{lung cancer}}
      ~
      \boxed{\includegraphics[width=0.8\textwidth,clip=True,trim=0 655 0 50]{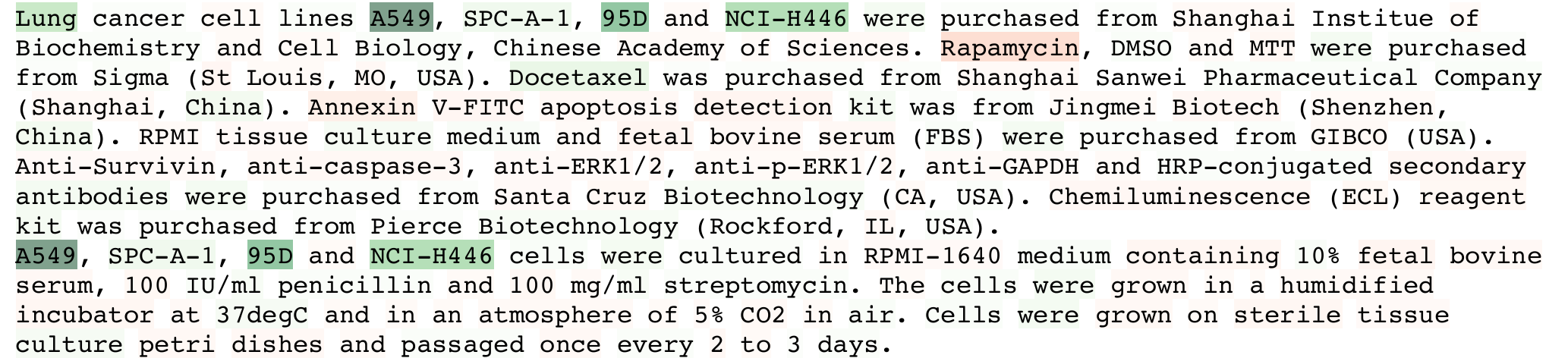}} \vskip 2mm
      \rotatebox{90}{\sf predicted \textbf{methods}}
      ~
      \boxed{\includegraphics[width=0.8\textwidth,clip=True,trim=0 655 0 50]{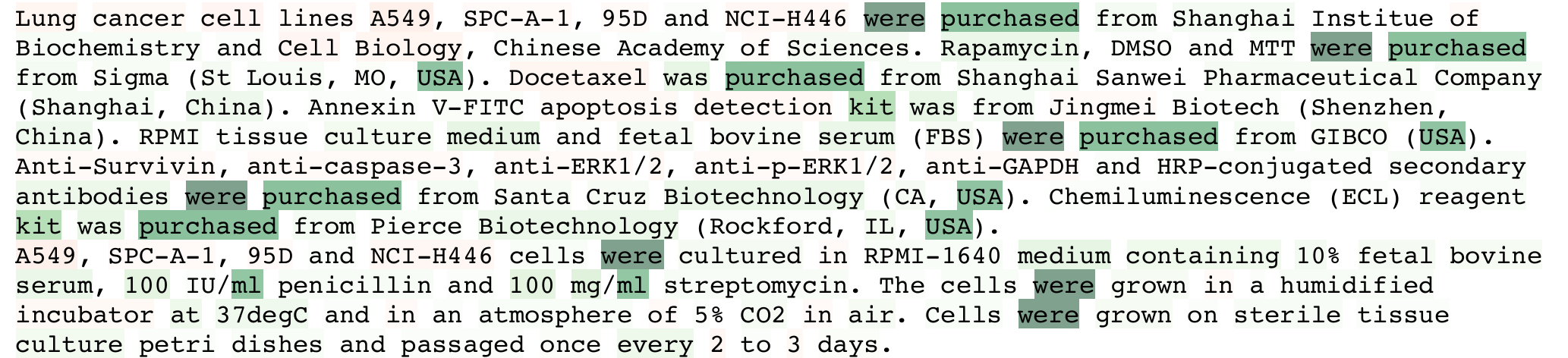}}
  \caption{Tf-idf features (\emph{top}) and LRP visualizations based on the classifier output from an SVM created for a paper's paragraph in the cancer type classification task (\emph{middle}) and the paragraph type classification task (\emph{bottom}).}
  \label{fig:highlighting_cancertype_partype}
\end{figure*}
When manually checking how often individual words occur in documents assigned to a specific class, it is no surprise to find that most paragraphs from papers about brain cancer contain the word `brain', while paragraphs from papers about breast cancer or prostate cancer often reference `woman'/`women' or `man'/`men' respectively (Fig.~\ref{fig:cancer_occurrences}). With classes named after different types of cancer, it is easy to guess which words might occur more often in some classes than others. However, when a dataset instead contains the classes `C1'-`C23', getting an idea of what is behind these labels would require manually looking at multiple documents of each class -- or extracting and visualizing relevant words for them.

We can observe subtle differences when examining the relevant words selected based on raw tf-idf features, using LRP on the classifier output of an SVM, and the scored distinctive words for all paragraphs of papers belonging to a single cancer type (Fig.~\ref{fig:wordcloud_cancertype}). For example, while tf-idf selected `patients' as a relevant word since it occurs frequently in some paragraphs yet still has only a moderate idf score, this obviously does not help to distinguish between different types of cancers. The relevant words selected by SVM+LRP as well as those ranked high when computing distinctive words seem more characteristic for this cancer type. However, due to the large tf-idf value, even with a positive but low classifier weight, `patients' is still ranked comparatively high when using SVM+LRP. SVM+LRP additionally provides negative scores, i.e.~it identifies words that contributed negatively to the classification decision, such as other cancer types like `breast cancer'.

The same analysis can be conducted taking the paragraph types as labels. When looking at the SVM+LRP word clouds for all five paragraph types (Fig.~\ref{fig:wordcloud_partype}), we can find many words and phrases typically used in these sections. For example, paragraphs from the results sections contain many references to tables and figures, while paragraphs in the discussion highlight the differences of `our study'. Further analysis could then, for example, reveal the differences in methodology when researching different types of cancer by extracting relevant words only from paragraphs belonging the methods sections, grouped by cancer type.

The classification accuracy of the SVM for cancer types ($95\%$) and paragraph types ($89.8\%$) is fairly good -- a prerequisite for using LRP in the first place. To better understand the classification process, we can create a heatmap visualization from the LRP scores of a single document. For example, when looking at a correctly classified paragraph from the methods section of a paper about lung cancer (Fig.~\ref{fig:highlighting_cancertype_partype}), we can clearly see how the classifier weights tuned for different tasks influence which words speak for or against either the corresponding cancer or paragraph type. Examining individual samples like this is particularly helpful to understand misclassifications and identify training set biases \cite{LapCVPR16}.

\subsection{New York Times articles}
It is especially interesting to see if the methods for selecting relevant words can help us identify trends in unlabeled datasets. For this we are using newspaper article snippets from the week of President Trump's inauguration (Jan $16^{\text{th}}$-$22^{\text{nd}}$, 2017), as well as three weeks prior (including the last week of 2016), downloaded with the Archive API from New York Times.\footnote{\url{https://developer.nytimes.com/archive_api.json}}
\begin{figure}[!h]
  \centering 
      \includegraphics[width=\columnwidth]{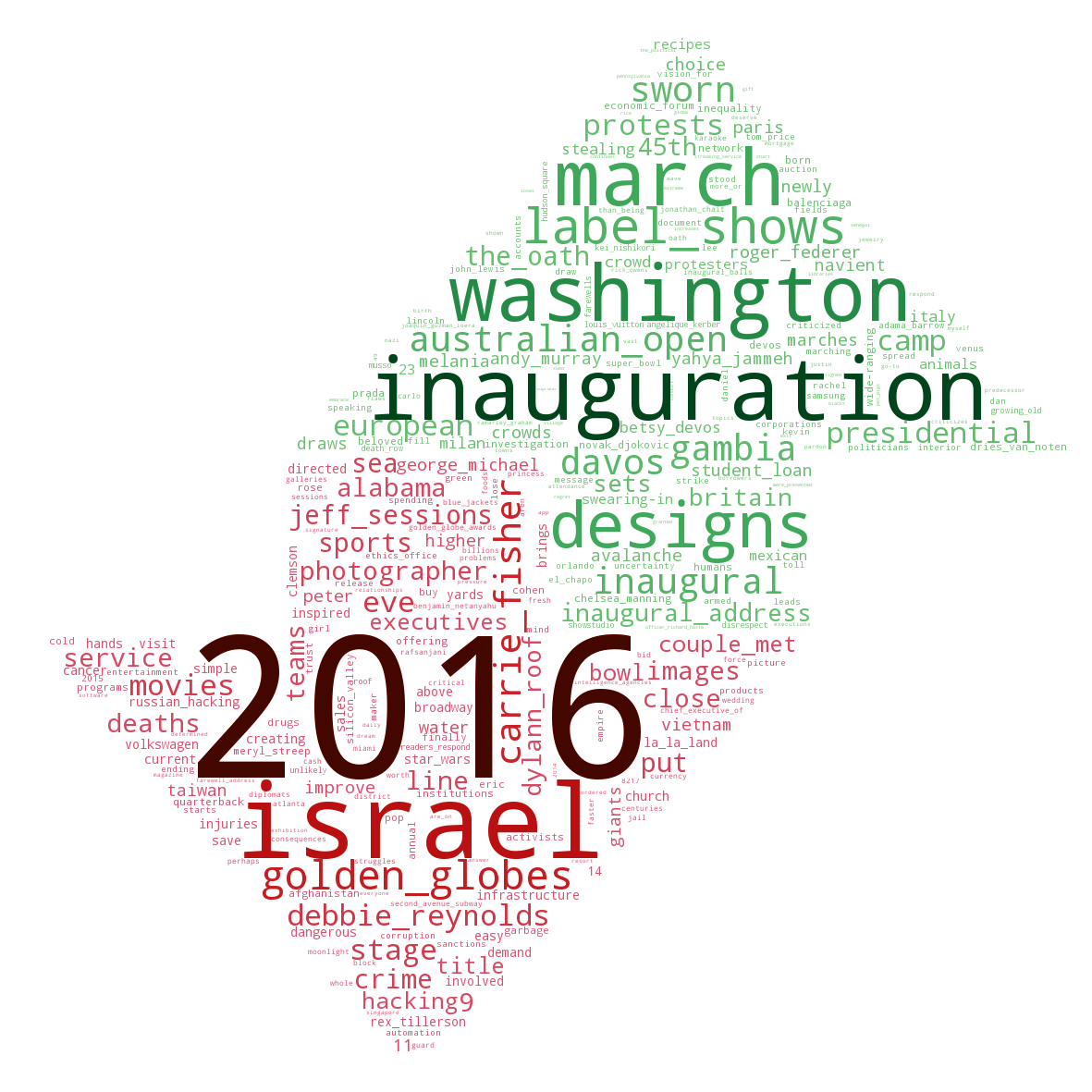}
  \caption{Distinctive words in NY Times article snippets during the week of president Trump's inauguration (\emph{green/up}) and three weeks prior (\emph{red/down}).}
  \label{fig:distinctive}
\end{figure}
\begin{figure}[!h]
  \centering 
      \includegraphics[width=\columnwidth]{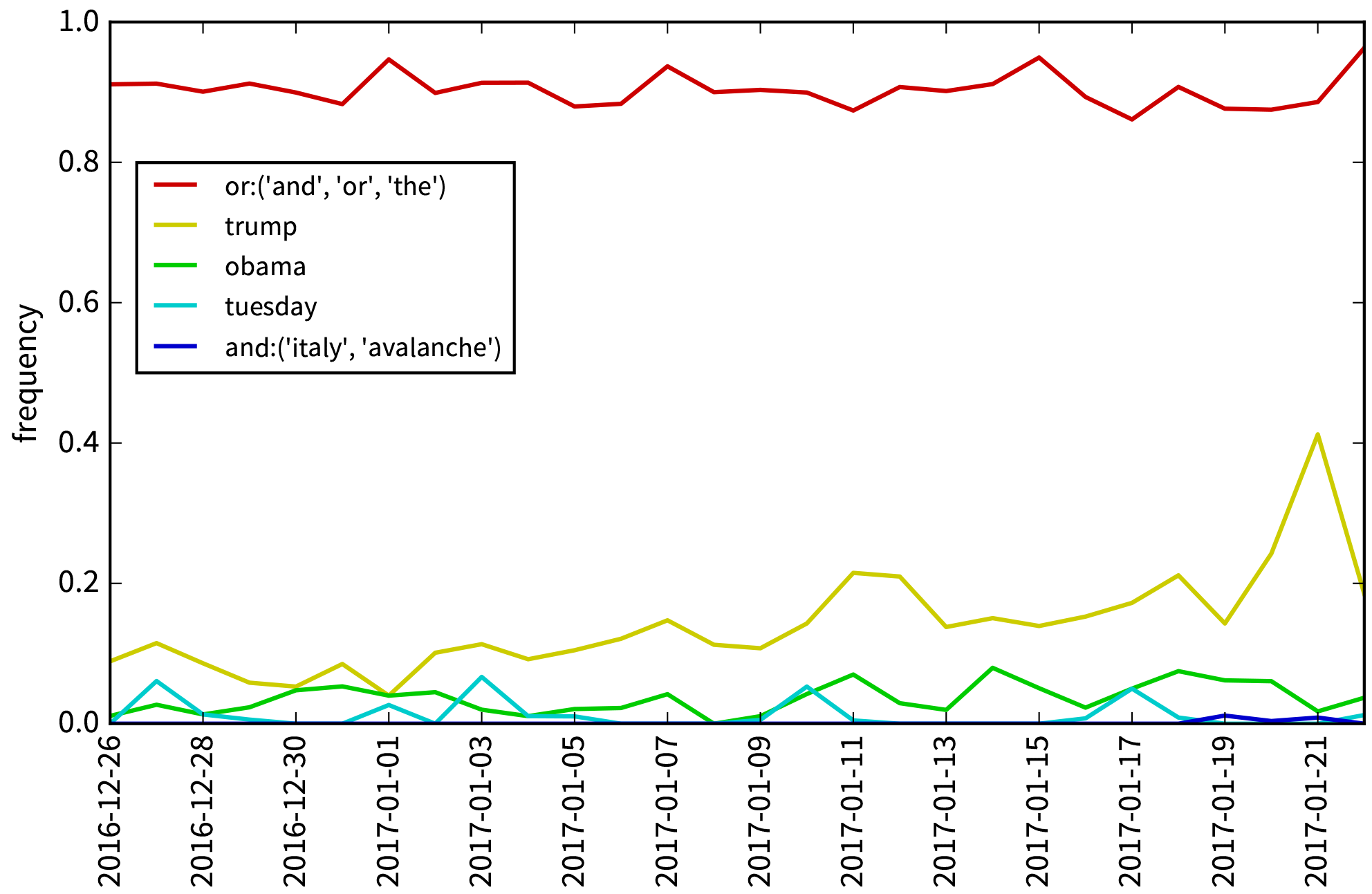}
  \caption{Frequencies of selected words in NY Times article snippets from different days.}
  \label{fig:nyt_occurrences}
\end{figure}
\begin{figure*}[!h]
  \centering 
      \includegraphics[width=0.3\textwidth]{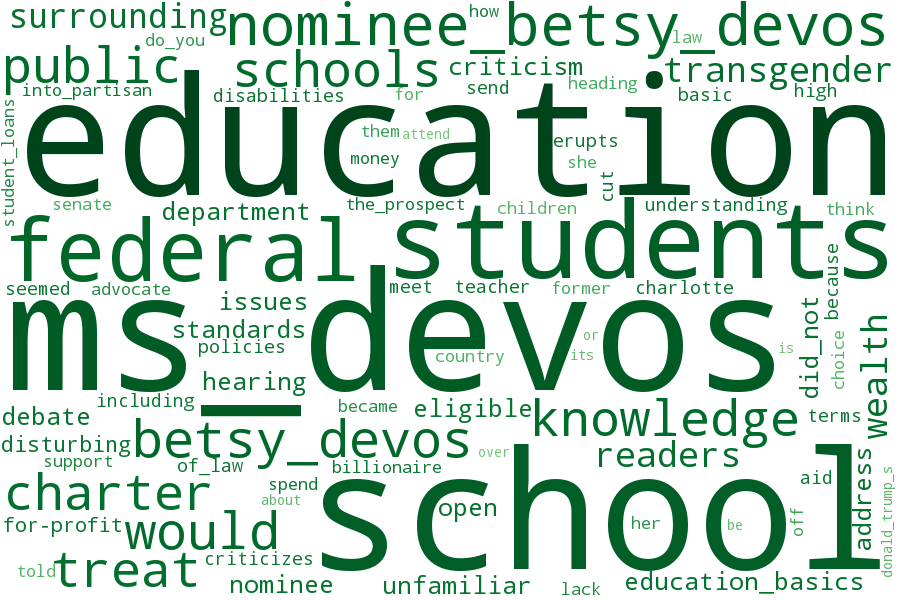}\hspace{0.32cm}
      \includegraphics[width=0.3\textwidth]{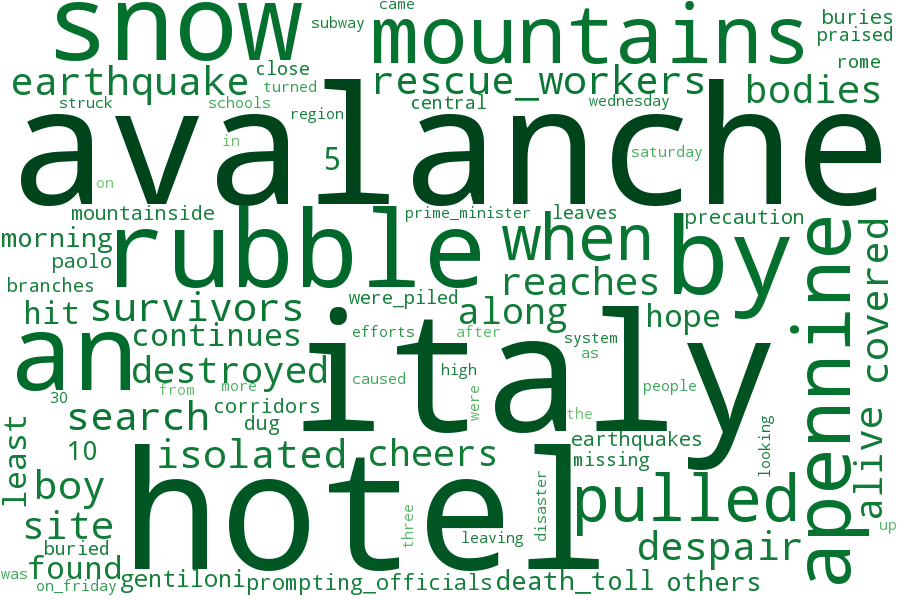}\\
      \includegraphics[width=0.65\textwidth]{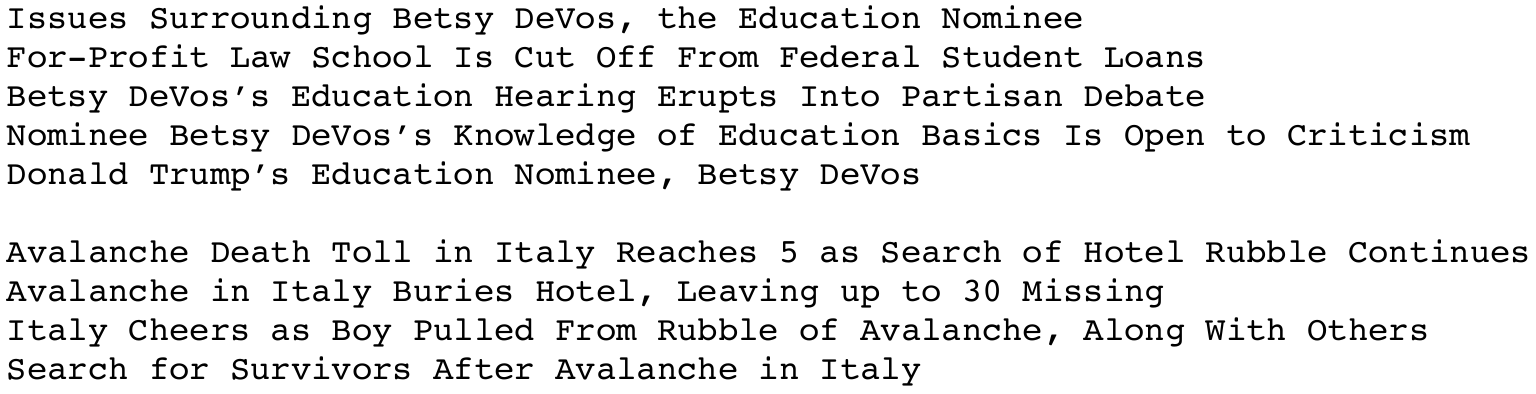}
  \caption{Word clouds created from the distinctive words identified for two of over 50 clusters during the week Jan $16^{\text{th}}$-$22^{\text{nd}}$, 2017 and corresponding headlines.}
  \label{fig:wordcloud_clusters}
\end{figure*}

When looking at the distinctive words identified for the week of the inauguration and before (Fig.~\ref{fig:distinctive}), we can see clear trends. Obviously, the main focus that week was on the inauguration itself, however it already becomes apparent that this will be followed by protest marches and also the Australian Open was happening at that time. When looking at the occurrence frequencies of different words over time (Fig.~\ref{fig:nyt_occurrences}), we can see the spike of `Trump' at the day of his inauguration, but while some stopwords occur equally frequent on all days, other rather meaningless words such as `Tuesday' have clear spikes as well (on Tuesdays). Therefore, care has to be taken when contrasting articles from different times when computing distinctive words, as it could easily happen that these meaningless words are picked up as well simply because e.g.~one month contains more Tuesdays than another month used for comparison.

To identify trending topics, the articles from the week of the inauguration were clustered using DBSCAN. When enforcing a minimum cosine similarity of $0.55$ to other samples of a cluster as well as at least three articles per cluster, we obtain over 50 clusters for this week (as well as several articles considered `noise'). While some clusters correspond to specific sections of the newspaper (e.g.~corrections to articles published the days before), others indeed refer to meaningful events that happened that week, e.g.~the nomination of Betsy DeVos or an avalanche in Italy (Fig.~\ref{fig:wordcloud_clusters}). 

\section{Conclusion}
Examining the relevant words occurring in documents of different classes or clusters in a dataset is a very helpful step in the exploratory analysis of a collection of texts. It allows to quickly grasp the contents of documents belonging to one class or cluster and can provide an intuition whether the proposed classification task is easy or difficult. Additionally, relevant words can help identify salient topics in unlabeled datasets, which is important if one is faced with a large dataset and quickly needs to find documents of interest.

We have explained three approaches for selecting relevant words of a class, namely 1)~aggregating the tf-idf scores of all texts belonging to this class, 2)~using layerwise relevance propagation to combine the weights of a trained classifier with the tf-idf feature vectors of this class, or 3)~directly computing a relevancy score for individual words depending on the number of documents in the target class this word occurs in compared to other classes. The usefulness of these methods was demonstrated by using the resulting word clouds to get an overview of the different classes in a dataset of scientific publications as well as to identify trending topics in recent New York Times article snippets. 

We found that for the exploratory analysis of different clusters and especially to identify trending topics, the best method for obtaining relevant words is to directly compute a score for each word indicating how often this word occurs in the current class compared to others. This method is very fast and robust with respect to varying numbers of samples per class. Most importantly, it is possible to identify distinctive words even if many clusters contain only very few samples, which would make it difficult to train a classifier on the dataset. Yet, when the documents are part of a classification task, it is very useful to use LRP to better understand the classification process and decisions.

We hope the provided code will encourage other people faced with large collections of texts to quickly dive into the analysis and to thoroughly explore new datasets.

\section*{Acknowledgments}
We would like to thank Ivana Balažević and Christoph Hartmann for their helpful comments on an earlier version of this manuscript. Franziska Horn acknowledges funding from the Elsa-Neumann scholarship from the TU Berlin.

\bibliography{../../phd_collected}

\begin{thebibliography}{20}
\providecommand{\natexlab}[1]{#1}
\providecommand{\url}[1]{\texttt{#1}}
\expandafter\ifx\csname urlstyle\endcsname\relax
  \providecommand{\doi}[1]{doi: #1}\else
  \providecommand{\doi}{doi: \begingroup \urlstyle{rm}\Url}\fi

\bibitem[Arras et~al.(2016{\natexlab{a}})Arras, Horn, Montavon, M{\"u}ller, and
  Samek]{arras2016explaining}
Leila Arras, Franziska Horn, Gr{\'e}goire Montavon, Klaus-Robert M{\"u}ller,
  and Wojciech Samek.
\newblock {Explaining Predictions of Non-Linear Classifiers in NLP}.
\newblock In \emph{Proceedings of the 1st Workshop on Representation Learning
  for NLP}, pages 1--7. Association for Computational Linguistics,
  2016{\natexlab{a}}.

\bibitem[Arras et~al.(2016{\natexlab{b}})Arras, Horn, Montavon, M{\"u}ller, and
  Samek]{arras2016relevant}
Leila Arras, Franziska Horn, Gr{\'e}goire Montavon, Klaus-Robert M{\"u}ller,
  and Wojciech Samek.
\newblock "what is relevant in a text document?": An interpretable machine
  learning approach.
\newblock \emph{arXiv preprint arXiv:1612.07843}, 2016{\natexlab{b}}.

\bibitem[Arras et~al.(2017)Arras, Montavon, M{\"u}ller, and
  Samek]{arras2017explaining}
Leila Arras, Gr{\'e}goire Montavon, Klaus-Robert M{\"u}ller, and Wojciech
  Samek.
\newblock Explaining recurrent neural network predictions in sentiment
  analysis.
\newblock \emph{arXiv preprint arXiv:1706.07206}, 2017.

\bibitem[Bach et~al.(2015)Bach, Binder, Montavon, Klauschen, M{\"u}ller, and
  Samek]{bach2015pixel}
Sebastian Bach, Alexander Binder, Gr{\'e}goire Montavon, Frederick Klauschen,
  Klaus-Robert M{\"u}ller, and Wojciech Samek.
\newblock On pixel-wise explanations for non-linear classifier decisions by
  layer-wise relevance propagation.
\newblock \emph{{PLOS ONE}}, 10\penalty0 (7):\penalty0 e0130140, 2015.

\bibitem[Ester et~al.(1996)Ester, Kriegel, Sander, Xu, et~al.]{ester1996dbscan}
Martin Ester, Hans-Peter Kriegel, J{\"o}rg Sander, Xiaowei Xu, et~al.
\newblock A density-based algorithm for discovering clusters in large spatial
  databases with noise.
\newblock In \emph{Kdd}, volume~96, pages 226--231, 1996.

\bibitem[Forman(2003)]{forman2003extensive}
George Forman.
\newblock An extensive empirical study of feature selection metrics for text
  classification.
\newblock \emph{The Journal of Machine Learning Research}, 3:\penalty0
  1289--1305, 2003.

\bibitem[Heimerl et~al.(2014)Heimerl, Lohmann, Lange, and
  Ertl]{heimerl2014word}
Florian Heimerl, Steffen Lohmann, Simon Lange, and Thomas Ertl.
\newblock Word cloud explorer: Text analytics based on word clouds.
\newblock In \emph{System Sciences (HICSS), 2014 47th Hawaii International
  Conference on}, pages 1833--1842. IEEE, 2014.

\bibitem[Hulth(2003)]{hulth2003improved}
Anette Hulth.
\newblock Improved automatic keyword extraction given more linguistic
  knowledge.
\newblock In \emph{Proceedings of the 2003 conference on Empirical methods in
  natural language processing}, pages 216--223. Association for Computational
  Linguistics, 2003.

\bibitem[Lapuschkin et~al.(2016)Lapuschkin, Binder, Montavon, M{\"u}ller, and
  Samek]{LapCVPR16}
Sebastian Lapuschkin, Alexander Binder, Gr{\'e}goire Montavon, Klaus-Robert
  M{\"u}ller, and Wojciech Samek.
\newblock Analyzing classifiers: Fisher vectors and deep neural networks.
\newblock In \emph{Proceedings of the IEEE Conference on Computer Vision and
  Pattern Recognition (CVPR)}, pages 2912--2920, 2016.

\bibitem[Lee and Kim(2008)]{lee2008news}
Sungjick Lee and Han-joon Kim.
\newblock News keyword extraction for topic tracking.
\newblock In \emph{Networked Computing and Advanced Information Management,
  2008. NCM'08. Fourth International Conference on}, volume~2, pages 554--559.
  IEEE, 2008.

\bibitem[Manning and Sch\"{u}tze(1999)]{fsnlp}
Christopher~D. Manning and Hinrich Sch\"{u}tze.
\newblock \emph{Foundations of Statistical Natural Language Processing}.
\newblock MIT Press, Cambridge, MA, USA, 1999.
\newblock ISBN 0-262-13360-1.

\bibitem[Manning et~al.(2008)Manning, Raghavan, and Sch\"{u}tze]{irbook}
Christopher~D. Manning, Prabhakar Raghavan, and Hinrich Sch\"{u}tze.
\newblock \emph{Introduction to Information Retrieval}.
\newblock Cambridge University Press, New York, NY, USA, 2008.
\newblock ISBN 0521865719, 9780521865715.

\bibitem[McNaught and Lam(2010)]{mcnaught2010using}
Carmel McNaught and Paul Lam.
\newblock Using wordle as a supplementary research tool.
\newblock \emph{The qualitative report}, 15\penalty0 (3):\penalty0 630, 2010.

\bibitem[Mikolov et~al.(2013)Mikolov, Sutskever, Chen, Corrado, and
  Dean]{mikolov2013distributed}
Tomas Mikolov, Ilya Sutskever, Kai Chen, Greg~S Corrado, and Jeff Dean.
\newblock Distributed representations of words and phrases and their
  compositionality.
\newblock In \emph{Advances in neural information processing systems}, pages
  3111--3119, 2013.

\bibitem[Montavon et~al.(2017)Montavon, Samek, and
  M{\"u}ller]{montavon2017methods}
Gr{\'e}goire Montavon, Wojciech Samek, and Klaus-Robert M{\"u}ller.
\newblock Methods for interpreting and understanding deep neural networks.
\newblock \emph{arXiv preprint arXiv:1706.07979}, 2017.

\bibitem[M{\"u}ller et~al.(2001)M{\"u}ller, Mika, R{\"a}tsch, Tsuda, and
  Sch{\"o}lkopf]{muller2001introduction}
Klaus-Robert M{\"u}ller, Sebastian Mika, Gunnar R{\"a}tsch, Koji Tsuda, and
  Bernhard Sch{\"o}lkopf.
\newblock An introduction to kernel-based learning algorithms.
\newblock \emph{Neural Networks, IEEE Transactions on}, 12\penalty0
  (2):\penalty0 181--201, 2001.

\bibitem[Sch{\"o}lkopf et~al.(1998)Sch{\"o}lkopf, Smola, and
  M{\"u}ller]{scholkopf1998nonlinear}
Bernhard Sch{\"o}lkopf, Alexander Smola, and Klaus-Robert M{\"u}ller.
\newblock Nonlinear component analysis as a kernel eigenvalue problem.
\newblock \emph{Neural computation}, 10\penalty0 (5):\penalty0 1299--1319,
  1998.

\bibitem[van~der Maaten and Hinton(2008)]{van2008visualizing}
Laurens van~der Maaten and Geoffrey Hinton.
\newblock {Visualizing data using t-SNE}.
\newblock \emph{Journal of Machine Learning Research}, 9\penalty0
  (2579-2605):\penalty0 85, 2008.

\bibitem[Yang and Pedersen(1997)]{yang1997comparative}
Yiming Yang and Jan~O. Pedersen.
\newblock A comparative study on feature selection in text categorization.
\newblock In \emph{Proceedings of the Fourteenth International Conference on
  Machine Learning}, ICML '97, pages 412--420, San Francisco, CA, USA, 1997.
  Morgan Kaufmann Publishers Inc.
\newblock ISBN 1-55860-486-3.

\bibitem[Zhang et~al.(2006)Zhang, Xu, Tang, and Li]{zhang2006keyword}
Kuo Zhang, Hui Xu, Jie Tang, and Juanzi Li.
\newblock \emph{Keyword Extraction Using Support Vector Machine}, pages 85--96.
\newblock Springer Berlin Heidelberg, Berlin, Heidelberg, 2006.

\end{thebibliography}
\bibliographystyle{plainnat}

\appendix
\section{Identifying bigrams}\label{sec:bigrams}
As an optional preprocessing step, single semantic units consisting of multiple words can be identified in the texts in order to preserve their special meaning. For example, the NBA team `Memphis Grizzlies' has nothing to do with actual grizzly bears and Google co-founder `Larry Page' is not like every other Larry. Therefore it would be desirable to identify these meaningful word combinations and not split them up into individual words, which, in a bag-of-words model, would cause them to loose their specific meaning. However, since we are later computing a relevancy score for each feature, we also don't want to consider just any combination of two or more words, which would lead to a computationally inefficient inflation of the feature space.

There are several ways to infer that a combination of two words (a so-called \emph{bigram}) constitutes a meaningful phrase~\cite{fsnlp}. In essence, these methods are based on a score, which is computed for every bigram that occurs in the texts, and if this score is above some threshold, those two words are subsequently considered as a single entity. For our purposes, a simple data-driven approach similar to~\cite{mikolov2013distributed} is adopted, where the score for every bigram consisting of the words $t_i$ and $t_j$ is computed as
\begin{align*}
\text{score}\,(t_i \, t_j) &= {\text{count}\,(t_i \, t_j) \over \max\{\text{count}\,(t_i),\, \text{count}\,(t_j)\}}.
\end{align*}
This score is equal to $1$ if both words occur only in this combination. 

Based on the computed scores, bigrams are identified in two steps: First, all bigrams, which do not occur at least twice in the texts, are discarded to ensure that the selected bigrams are not just a random combination of infrequent words. Next, it is checked that the score is above a decision threshold to separate true bigrams from coincidental co-occurrences of two words.

\begin{figure}[!h]
  \centering
      \includegraphics[width=\columnwidth]{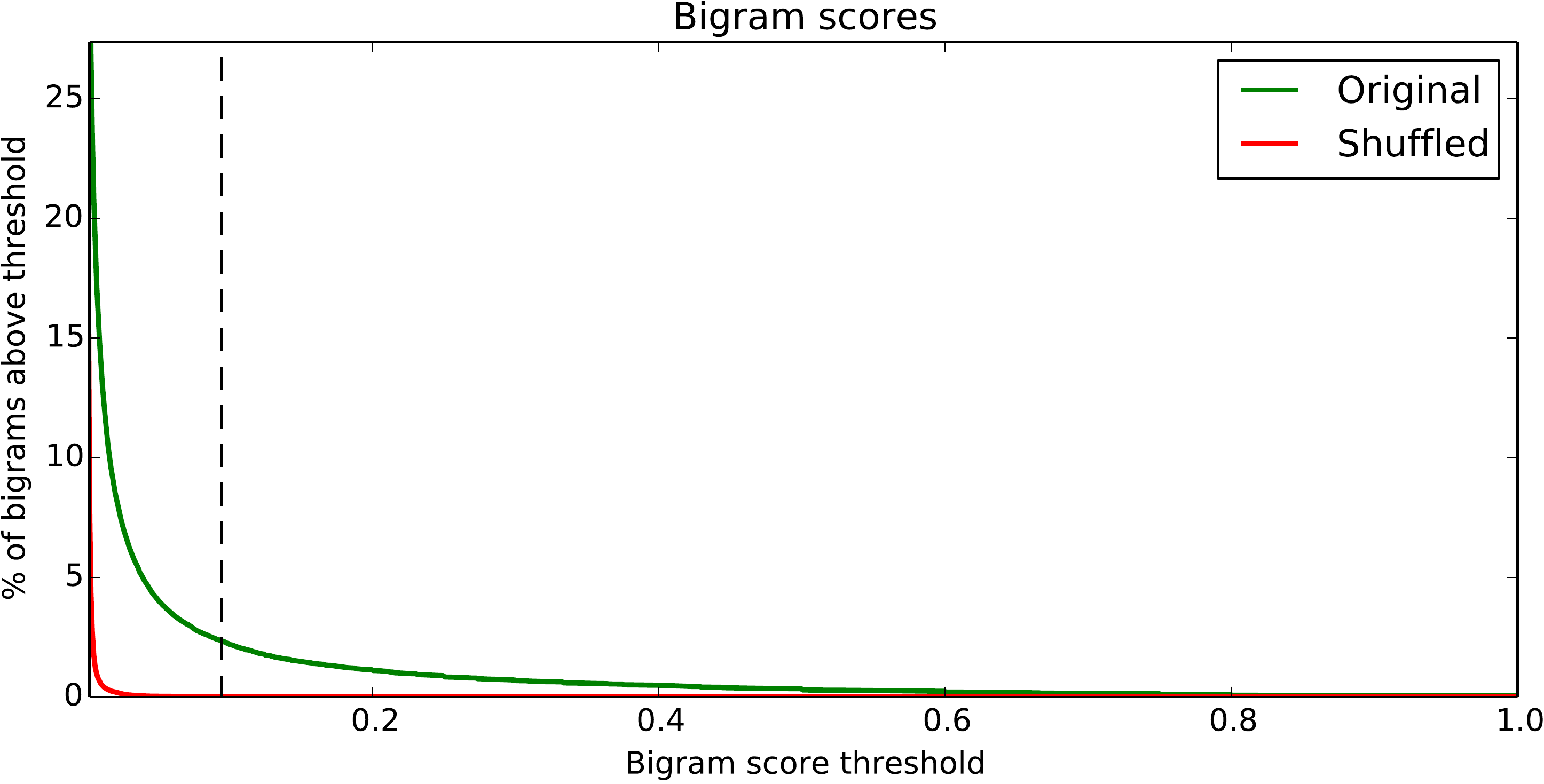}
  \caption{Percentage of the bigrams in original texts and documents with randomly shuffled words whose scores would pass certain thresholds.}
  \label{fig:bigramdist}
\end{figure}

In order to choose this threshold in an objective manner, an empirical hypothesis approach is adopted. To this end, the distribution of bigram scores from original texts is compared to that obtained from documents with randomly permuted words.\footnote{To obtain the random scores, the words were only shuffled within a document, not across documents.}
A large majority of random bigrams have a score around zero, therefore when choosing a very conservative threshold such as $0.1$, $99.99\%$ of the random bigrams' scores are below this threshold and meaningful word combinations can be selected very reliably (Fig.~\ref{fig:bigramdist}). 

When joining all bigrams, whose scores pass the threshold, together as a single phrase, this can actually lead to longer word combinations than just bigrams; e.g.~if `President Barack Obama' is a common phrase, both `President Barack' and `Barack Obama' will have a high bigram score and will therefore be joined together, yielding the trigram `President Barack Obama'.

\end{document}